\documentclass[letterpaper]{article} 
\usepackage{aaai2027}  
\usepackage[hyphens]{url}  
\usepackage{graphicx} 
\usepackage{natbib}  
\usepackage{caption} 
\usepackage{algorithm}
\usepackage{algorithmic}
\usepackage{multirow}
\usepackage{amsmath}
\usepackage{amssymb}
\usepackage{newfloat}
\usepackage{listings}
\DeclareCaptionStyle{ruled}{labelfont=normalfont,labelsep=colon,strut=off} 
\floatstyle{ruled}
\newfloat{listing}{tb}{lst}{}
\floatname{listing}{Listing}

\usepackage{booktabs}
\usepackage{xspace}
\newcommand{\paratitle}[1]{\vspace{1.5ex}\noindent\textbf{#1}}
\newcommand{\ourmodel}{TurnSight\xspace}
\newcommand{\wo}{\emph{w/o}\xspace}
\title{TurnSight: Turn-Level Hindsight Self-Distillation for Tool-Integrated Reasoning}

\author{
Changle Qu\textsuperscript{\rm 1},
Sunhao Dai\textsuperscript{\rm 1},
Hengyi Cai\textsuperscript{\rm 2},
Yuqi Zhou\textsuperscript{\rm 1},
Xinran Chen\textsuperscript{\rm 2},
Simon\textsuperscript{\rm 2},
Jun Xu\textsuperscript{\rm 1}
}

\affiliations{
\textsuperscript{\rm 1}Gaoling School of Artificial Intelligence, Renmin University of China\
\textsuperscript{\rm 2}Baidu Inc.\\
\{changlequ,junxu\}@ruc.edu.cn
}

\begin{document}

\maketitle

\begin{abstract}
Tool-Integrated Reasoning (TIR) enables LLMs to solve complex tasks through iterative tool interactions. 
However, existing reinforcement learning methods often rely on trajectory-level supervision, limiting fine-grained credit assignment in long-horizon TIR scenarios.
On-policy self-distillation offers denser signals through teacher branches with privileged context, but existing approaches typically derive such context from ground-truth answers or retrieved skills, which may not reflect the states actually visited by the agent. 
Moreover, token-level supervision fails to capture the turn-level structure of tool interactions.
To address this, we propose \ourmodel, a turn-level hindsight self-distillation framework that derives supervision directly from execution-conditioned hindsight. 
It then constructs multiple hindsight views with different lookahead horizons and selects reliable supervision through cross-horizon directional agreement.
Finally, the selected hindsight signal is normalized across sibling rollouts and used to adaptively modulate RL advantages while preserving their original optimization direction.
Extensive experiments on three benchmarks demonstrate the effectiveness of \ourmodel.
Our codes are available at~\textcolor{blue}{\url{https://github.com/quchangle1/TurnSight}}. 

\end{abstract}

\section{Introduction}
Tool-Integrated Reasoning (TIR) enables large language models (LLMs) to solve complex tasks by interleaving reasoning with external tool interactions~\cite{gou2023tora,qu2025from,qu2025tool}.
Unlike single-step generation, a TIR agent repeatedly selects tools, constructs arguments, observes execution results, and adapts its subsequent behavior.
These interactions substantially expand the capabilities of LLMs, but also create a difficult temporal credit-assignment problem: the final task outcome reveals whether a trajectory succeeded, yet provides little information about which intermediate tool interactions are useful, redundant, or harmful.

\begin{figure}[t]
\centering
	\includegraphics[width=\linewidth]{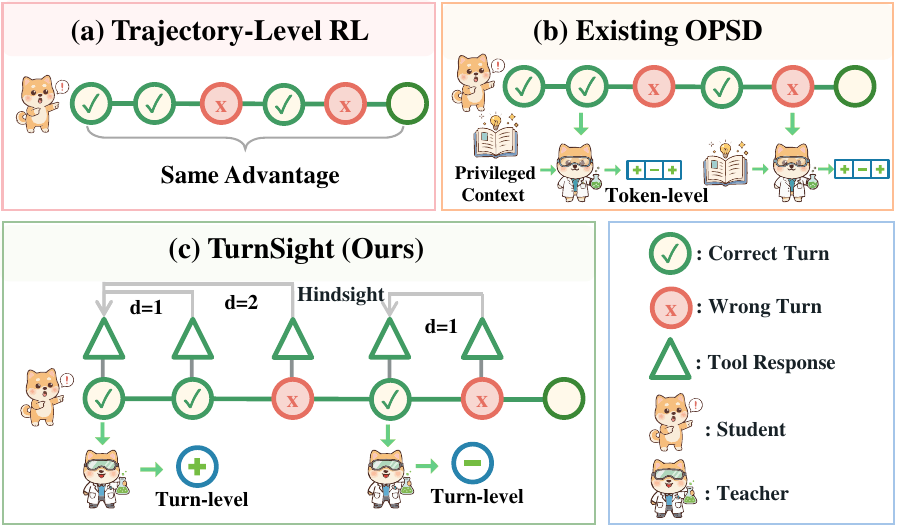}
\caption{Comparison of credit assignment strategies for TIR. (a) Trajectory-level RL assigns identical credit to all turns. (b) Existing OPSD may use state-misaligned privileged context and produce conflicting token-level signals. (c) TurnSight derives coherent turn-level credit from execution-conditioned hindsight across multiple lookahead depths.}
        \label{fig:intro}
        \vspace{-0.2cm}
\end{figure}

Reinforcement learning with verifiable rewards (RLVR) provides a natural framework for training TIR agents~\cite{chang2025thor,jiang2025verltool}.
However, as shown in Figure~\ref{fig:intro}(a), outcome-based methods such as GRPO typically propagate a single trajectory-level advantage to all generated actions, assigning similar credit to decisions with very different causal effects~\cite{shao2024deepseekmath,zeng2025tool}.
Recent methods provide denser supervision by matching generated tool calls against annotated or reference tool trajectories~\cite{qian2025toolrl,qu2026matchtir}.
Although effective, such supervision assumes that the reference action remains appropriate at the state visited by the learned policy.
This assumption becomes fragile in multi-turn interaction: once the agent makes a different tool call, the environment changes, and subsequent states may diverge from the reference trajectory. 
Thus, effective credit assignment for TIR requires supervision that is both fine-grained and state-aligned.


On-policy self-distillation (OPSD)~\cite{shenfeld2026sdft,li2026rethinking,kim2026does} offers a promising way to obtain such supervision. 
It evaluates student-generated behavior using a privileged view available only during training, thereby providing feedback on on-policy states rather than prescribing an external trajectory. 
However, applying OPSD to multi-turn TIR is nevertheless non-trivial. 
As shown in Figure~\ref{fig:intro}(b), existing methods often enrich the teacher with globally informative context, such as ground-truth answers, reference rollouts, or distilled skills~\cite{zhao2026opsd,hubotter2026sdpo,lu2026sdar,ye2026opcd}.
These signals describe how to solve the task, but not whether a tool interaction was appropriate at the state where it occurred. 
This mismatch is especially pronounced in TIR, where each tool call changes the external environment and may cause subsequent states to diverge from the reference path.
In this setting, the trajectory itself contains the most state-aligned supervision, with tool execution outcomes serving as hindsight signals.

State alignment alone is insufficient, however. Existing self-distillation methods generally produce token-level signals, whereas the fundamental decision unit in TIR is a complete tool interaction, comprising reasoning, tool selection, and argument construction.
Within a turn, token-level supervision can vary in magnitude or even sign due to formatting tokens and argument-level variations.
Optimizing these signals independently can therefore assign conflicting credit within the same interaction.
This motivates our turn-level formulation, which aggregates hindsight supervision at the interaction level to provide more coherent  credit assignment.

In this paper, we propose \ourmodel, a turn-level hindsight self-distillation method for multi-turn TIR.
Instead of relying on external reference trajectories, \ourmodel constructs hindsight supervision directly from on-policy tool execution outcomes.
To better match the decision granularity of multi-turn TIR, it first aggregates token-level hindsight evidence into coherent turn-level signals.
It then constructs multiple hindsight views with different lookahead horizons and selects reliable supervision through cross-horizon directional agreement. 
Finally, the selected hindsight signal is normalized across sibling rollouts and used to construct bounded, sign-aware weights that modulate RL advantages without changing their optimization direction.
As a result, \ourmodel introduces no auxiliary imitation objective, is independent of the underlying reward construction, and can be seamlessly integrated with existing policy-gradient RL algorithms.

Extensive experiments on both in-domain and out-of-domain benchmarks verify the effectiveness of \ourmodel and its ability to generalize across diverse tool-use scenarios.
In summary, our main contributions are as follows:


$\bullet$ We formulate two requirements for credit assignment in multi-turn TIR: supervision should be aligned with on-policy execution states and coherent at the interaction level.

$\bullet$ We propose \ourmodel, which derives execution-conditioned, multi-horizon hindsight assessments to modulate RL advantages without changing their direction.

$\bullet$ Extensive experiments on both in-domain and out-of-domain benchmarks demonstrate the effectiveness, robustness, and generalization ability of \ourmodel.

\section{Related Work}
\subsection{Tool-Integrated Reasoning}
Recent RLVR methods, particularly GRPO-style algorithms \cite{shao2024deepseekmath,yu2025dapo}, have emerged as an effective paradigm for improving TIR by directly optimizing models with task-level feedback \cite{wang2025otc,xue2025simpletir,qu2025tool}. 
Early approaches typically rely on sparse outcome rewards \cite{li2025torl,feng2025retool} or trajectory-level supervision \cite{qian2025toolrl,zhang2025nemotron,zeng2025tool,wei2025autotir}, where all reasoning turns within a trajectory receive identical advantage signals, limiting fine-grained credit assignment. 
Recent efforts improve this by introducing turn-level supervision through LLM-based evaluation \cite{wang2025information} or by comparing predicted tool calls with ground-truth tool executions~\cite{qu2026matchtir}. 
However, these methods rely on annotated or reference tool trajectories, making optimization closer to imitation learning and limiting exploration. In contrast, \ourmodel combines RL with on-policy self-distillation to achieve fine-grained credit assignment without ground-truth trajectories while preserving exploration.

\subsection{On-Policy Self-Distillation}

On-policy self-distillation evaluates student-generated trajectories using a privileged teacher view constructed from the same model with additional information, providing dense supervision on the states visited by the student~\citep{sang2026crisp,ma2026sdsearch,li2026unifying,wang2026skill}.
Rather than directly matching the teacher distribution, recent methods use teacher--student disagreement to reweight RL advantages, allocate fine-grained credit, or bound distillation interventions \citep{yang2026rlsd,zhang2026stepopsd,ding2026sgcd,zhou2026turnopd,zhou2026sageopd}. 
Despite these advances, existing methods often rely on globally informative teacher signals and token-level distillation, which may not provide state-aligned supervision at the natural turn-level granularity of TIR.
In contrast, we propose \ourmodel, which constructs turn-level hindsight supervision directly from on-policy tool execution outcomes, enabling fine-grained and state-aligned credit assignment.

\begin{figure*}[t]
\centering
	\includegraphics[width=\linewidth]{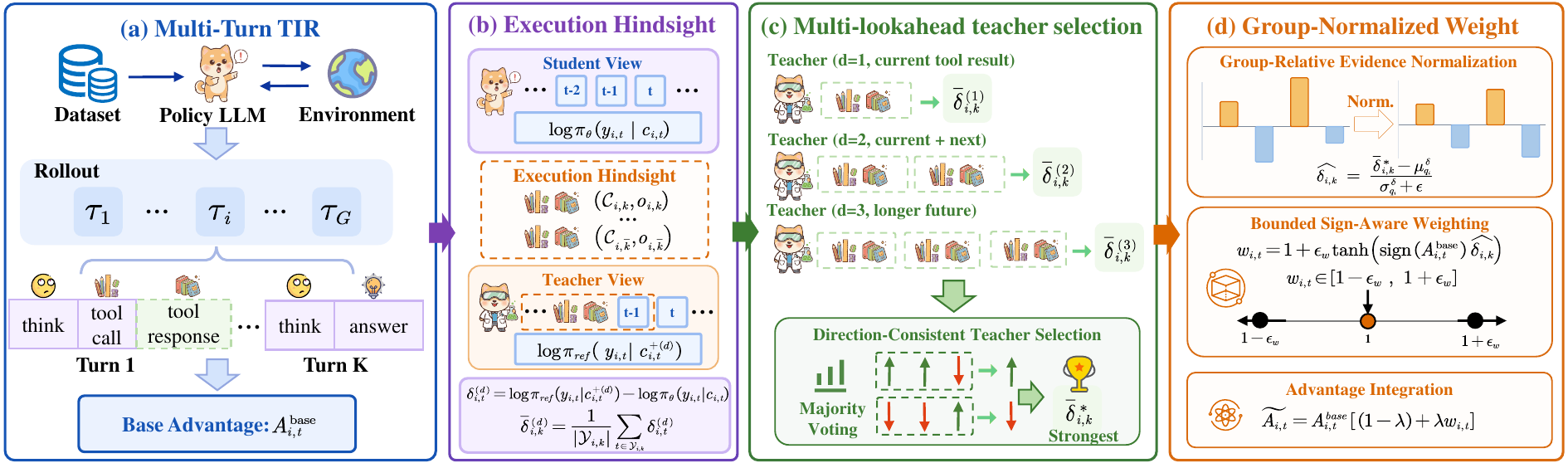}
\caption{The illustration of our proposed turn-level hindsight self-distillation framework \ourmodel for TIR.}
        \label{fig:method}
\end{figure*}

\section{Preliminaries}
In this section, we first formulate the task of TIR, followed by the agentic RL framework and the OPSD mechanism.

\subsection{Task Formulation}

Given a user prompt $q \sim \mathcal{D}$ and a set of $M$ tools $\mathcal{T}=\{t_1,\ldots,t_M\}$, where each tool is specified by its name and argument schema, an agent with policy $\pi_\theta$ interacts with an environment to produce a trajectory $\tau=(s_1,\ldots,s_K)$. 
We represent a non-terminal turn as $s_k=(n_k,\mathcal{C}_k,o_k)$, where $n_k$ is the natural-language reasoning and $\mathcal{C}_k$ is the set of tool calls issued in that turn. Each call contains a tool name and a JSON argument object. 
A trajectory terminates when the policy emits its final answer or reaches the pre-defined maximum turn limit $L$. For the terminal turn, $\mathcal{C}_K=\emptyset$ and $o_K=\emptyset$. 
Our goal is to achieve fine-grained turn-level credit assignment by leveraging execution-aware supervision.

\subsection{Agentic Reinforcement Learning}
We optimize the agent policy $\pi_{\theta}$ using GRPO~\cite{shao2024deepseekmath}. For each prompt 
$q \sim \mathcal{D}$, the behavior policy $\pi_{\theta_{\mathrm{old}}}$ 
samples a group of $G$ interaction trajectories 
$\{\tau_i\}_{i=1}^{G}$. Rather than training an additional critic to 
predict state values, GRPO constructs policy-gradient advantages by 
comparing the rewards of trajectories generated for the same prompt. 
This group-wise comparison provides a prompt-dependent baseline and 
reduces the cost of value-function estimation.
The optimization objective 
is written as
\begin{equation*}
    \begin{split}
        &\mathcal{J}_{\text{GRPO}}(\theta)=\mathbb{E}_{q\sim D,\{\tau_i\}\sim\pi_{\theta_\text{old}}(\cdot|q)}\frac{1}{G}\sum_{i=1}^{G}\frac{1}{|\tau_i|}\sum_{t=1}^{|\tau_i|}\bigg[\min(\\
        &\rho_{i,t}\hat{A}_{i,t},\text{clip}(\rho_{i,t},
        1-\epsilon,1+\epsilon)\hat{A}_{i,t})-\beta \mathbb{D}_\text{KL}(\pi_{\theta}|\pi_\text{ref})\bigg],
    \end{split}
\label{eq:grpo}
\end{equation*}
where $\rho_{i,t} = \frac{\pi_{\theta}(\tau_{i,t}|q, \tau_{i,<t})}{\pi_{\theta_{\text{old}}}(\tau_{i,t}|q, \tau_{i,<t})}$ is the importance sampling ratio, $\widehat{A}_{i,t}$ is the normalized group-relative advantage, $\epsilon$ is the clipping parameter, and $\beta$ controls the KL regularization toward the reference policy $\pi_{\mathrm{ref}}$. 
Since trajectories contain both model-generated actions and environment observations, only tokens generated by the policy are optimized, while observation tokens are excluded from gradient computation.

\subsection{On-Policy Self-Distillation}

OPSD scores a trajectory sampled by the student under two information conditions. At token position $t$, the student distribution $p_t^{S}=\pi_\theta(\cdot\mid c_t)$ uses only rollout information, while the detached teacher distribution $p_t^{T}=\pi_T(\cdot\mid c_t^{+})$ additionally accesses privileged training-time information. For a sampled token $y_t$, we define the log-probability gap as
\begin{equation}
\delta_t
=
\log \pi_T(y_t\mid c_t^{+})
-
\log \pi_\theta(y_t\mid c_t).
\label{eq:token_gap}
\end{equation}
The gap $\delta_t$ captures how privileged information changes confidence in the realized token. 
Unlike direct distribution matching, which lets the teacher determine the update direction, our method uses $\delta_t$ as hindsight evidence that modulates RL advantage magnitudes while preserving their directions.

\section{Our Approach: \ourmodel}
In this section, we will provide a detailed introduction to our \ourmodel. The overall framework is illustrated in Figure~\ref{fig:method}.

\subsection{Execution-Conditioned Hindsight Construction}
\paratitle{Execution-Conditioned Privileged Context.}
In multi-turn TIR, each tool execution changes the environment and affects subsequent states.
Existing privileged context sources, such as ground-truth answers and reference rollouts, are not conditioned on the student's trajectory and may provide misaligned supervision.
We therefore construct privileged context directly from the execution outcomes generated by the student's own trajectory, which are naturally aligned with its visited states and provide informative hindsight for subsequent policy optimization.
Specifically, consider a token $y_{i,t}$ in turn $k$ of trajectory $\tau_i$. The student scores this token under its original causal context $c_{i,t}$, which contains only the information available during rollout. For a lookahead depth $d$, we collect the execution hindsight block
\[
\mathcal{H}_{i,k}^{(d)}
=
\Big(
(\mathcal{C}_{i,k},o_{i,k}),
\ldots,
(\mathcal{C}_{i,\bar{k}},o_{i,\bar{k}})
\Big),
\]
where $\mathcal{C}_{i,j}$ denotes the tool calls issued by the student in turn $j$, $o_{i,j}$ denotes the corresponding tool responses, and $\bar{k}=\min(k+d-1,K_i)$. So the privileged context is
\[
c_{i,t}^{+(d)}
=
c_{i,t}\oplus\mathcal{H}_{i,k}^{(d)},
\]
where the hindsight block is used only during training while the reference branch teacher-forces the original student response. We instantiate this branch with a frozen initial reference policy $\pi_{\mathrm{ref}}$. For each lookahead depth $d\in\mathcal{D}_H$, we compute the detached sampled-token gap
\[
\delta_{i,t}^{(d)}
=
\log\pi_{\mathrm{ref}}(y_{i,t}\mid c_{i,t}^{+(d)})
-
\log\pi_\theta(y_{i,t}\mid c_{i,t}).
\]
A positive gap indicates that observing the execution consequences increases the reference branch's confidence in the realized token, whereas a negative gap indicates the opposite.

\paratitle{Turn-Level Aggregation.}
Although hindsight gaps are computed at the token level, the natural decision unit in multi-turn TIR is an interaction turn, where reasoning, tool selection, and argument generation together constitute one tool-use decision. 
In practice, token-level hindsight signals within the same turn often vary substantially or even disagree due to formatting tokens and argument variations. 
Applying these signals independently may therefore produce inconsistent credit assignment.
To preserve the semantic coherence of each interaction, we aggregate token-level hindsight into a turn-level signal.
Let $\mathcal{Y}_{i,k}=\{t:\kappa_i(t)=k\}$ denote the token positions belonging to turn $k$. For each lookahead depth, we compute
\[
\bar{\delta}_{i,k}^{(d)}
=
\frac{1}{|\mathcal{Y}_{i,k}|}
\sum_{t\in\mathcal{Y}_{i,k}}
\delta_{i,t}^{(d)}.
\]
The resulting $\bar{\delta}_{i,k}^{(d)}$ provides a single hindsight assessment for the entire interaction turn, which is subsequently shared by all policy tokens in that turn and serves as the basic evidence unit for multi-lookahead teacher selection.

\subsection{Multi-Lookahead Teacher Selection}
\paratitle{Multi-Lookahead Hindsight Views.}
The amount of future information required to evaluate a tool-use decision varies across interaction states in multi-turn TIR. 
Some decisions, such as validating tool selection or argument correctness, can be assessed immediately from execution feedback, while others require subsequent tool interactions to reveal their utility.
Consequently, no single hindsight horizon is suitable for every interaction state: short horizons may overlook delayed effects, while long horizons may introduce unrelated future interactions and dilute the evidence for the current decision.

To capture these diverse temporal dependencies, we construct multiple execution-conditioned hindsight contexts with lookahead depths $\mathcal{D}_H=\{1,2,3\}$. 
Each lookahead depth instantiates an independent hindsight teacher by exposing the frozen reference policy to different amounts of future execution information, yielding a turn-level hindsight signal $\bar{\delta}_{i,k}^{(d)}$. 
Collectively, these teachers provide complementary assessments of the same tool-use decision, with shallow horizons emphasizing immediate execution quality and deeper ones reflecting downstream trajectory contributions.

\paratitle{Direction-Consistent Teacher Selection.}
Since teachers observe different amounts of future information, their hindsight assessments may disagree for the same interaction turn. Treating all teachers equally may weaken informative supervision, while selecting the strongest teacher alone may amplify an isolated and unreliable assessment. We therefore perform teacher selection in two stages: we first determine the reliable supervision direction through cross-horizon agreement and then select the strongest teacher within that direction.
For each turn, we first map teacher signal to its direction
\[
s_{i,k}^{(d)}
=
\begin{cases}
+1, & \bar{\delta}_{i,k}^{(d)} \geq 0,\\
-1, & \bar{\delta}_{i,k}^{(d)} < 0.
\end{cases}
\]

We then determine the consensus direction by majority voting over $\{s_{i,k}^{(d)}\}_{d\in\mathcal{D}_H}$ and select the teacher with the largest hindsight signal among those consistent with the consensus:
\[
d_{i,k}^{*}
=
\arg\max_{\substack{d\in\mathcal{D}_H:\\
s_{i,k}^{(d)}=v_{i,k}}}
\left|\bar{\delta}_{i,k}^{(d)}\right|,
\qquad
\bar{\delta}_{i,k}^{*}
=
\bar{\delta}_{i,k}^{(d_{i,k}^{*})},
\]
where $v_{i,k}$ denotes the majority direction.

Directional agreement filters out isolated teachers whose assessments conflict with the majority of hindsight views, while selecting the strongest consistent teacher preserves the most informative supervision without diluting it through indiscriminate fusion. The $\bar{\delta}_{i,k}^{*}$ is used as the final hindsight signal and is subsequently passed to group normalization.

\subsection{Group-Normalized Weight Construction}

\paratitle{Group-Relative Evidence Normalization.}
Although the selected hindsight signal $\bar{\delta}_{i,k}^{*}$ provides a turn-level assessment of each tool interaction, its magnitude is not directly comparable across trajectories. The teacher--student gap is affected by factors such as prompt difficulty, trajectory length, turn position, and training progress. Consequently, directly using the raw gap may overemphasize naturally large-scale examples instead of reflecting their relative hindsight evidence.

Inspired by GRPO, we normalize the selected hindsight evidence within each prompt group. Instead of grouping interactions by turn index, we compute a shared mean $\mu_{q_i}^{\delta}$ and standard deviation $\sigma_{q_i}^{\delta}$ over the selected token-level gaps from all valid turns and policy tokens across all sibling rollouts sampled for query $q_i$. The normalized signal is
\[
\widehat{\delta}_{i,k}
=
\frac{
\bar{\delta}_{i,k}^{*}-\mu_{q_i}^{\delta}
}{
\sigma_{q_i}^{\delta}+\epsilon
}.
\]

The normalized signal measures the relative hindsight evidence of the current turn against all valid interactions across sibling on-policy rollouts for the same query, reducing prompt-specific scale variation and making the signal more comparable across queries and throughout training.

\paratitle{Bounded Sign-Aware Weighting.}
The normalized hindsight signal is used to modulate the RL credit assigned by the base advantage. We therefore construct a bounded weight
\[
w_{i,t}
=
1+\epsilon_w
\tanh\!\left(
\mathrm{sign}\!\left(A_{i,t}^{\mathrm{base}}\right)
\widehat{\delta}_{i,k}
\right).
\]
The sign term aligns hindsight evidence with the update direction determined by the base advantage. Consequently, agreement between hindsight and the RL signal increases the update magnitude ($w_{i,t}>1$), whereas disagreement attenuates it ($w_{i,t}<1$).
To mitigate the influence of extreme hindsight signals, we further introduce a bounded modulation mechanism, where $\epsilon_w$ controls the adjustment range:
\[
w_{i,t}\in[1-\epsilon_w,\,1+\epsilon_w].
\]

\begin{table*}[t]
    \centering
    \resizebox{1.0\textwidth}{!}{
    \begin{tabular}{l|cccc|ccccccc|c|c}
        \toprule
        \multirow{3}{*}{\textbf{Methods}}
        & \multicolumn{4}{c|}{\textbf{FTRL}}
        & \multicolumn{7}{c|}{\textbf{BFCL}}
        & \multicolumn{1}{c|}{\textbf{ToolHop}}
        & \multirow{3}{*}{\textbf{Avg.}} \\
        \cmidrule(lr){2-5}
        \cmidrule(lr){6-12}
        \cmidrule(lr){13-13}
        
        & \multirow{2}{*}{Solve-P}
        & \multirow{2}{*}{Solve-R}
        & \multirow{2}{*}{Solve-F1}
        & \multirow{2}{*}{Avg.}
        & \multicolumn{4}{c}{Multi-Turn}
        & \multicolumn{2}{c}{Agentic}
        & \multirow{2}{*}{Avg.} & \multirow{2}{*}{Acc.} \\
        \cmidrule(lr){6-9}
        \cmidrule(lr){10-11}
        
        & & & &
        & Base & MF & MP & LC
        & Search & Memory &
        & & \\
        \hline
        \multicolumn{14}{c}{\rule[-0.7ex]{0pt}{3.2ex}\textit{\textbf{Qwen3-4B}}} \\
        \midrule
        Vanilla
        & 30.78 & 29.65 & 25.85 & 28.76
        & 37.00 & 36.00 & 31.50 & 22.50
        & 9.00 & 19.14 & 22.91
        & 23.22
        & 24.96 \\
        
        GRPO
        & 31.61 & 32.64 & 28.49 & 30.91
        & 42.00 & 34.00 & 29.00 & 23.50
        & 8.00 & 19.35 & 22.90
        & 21.98
        & 25.26 \\
        
        ToolRL
        & 34.07 & 31.63 & 27.94 & 31.21
        & 43.00 & 41.00 & 27.50 & 25.50
        & 8.50 & \underline{24.09} & 25.27
        & 23.65
        & 26.71 \\
        
        MatchTIR
        & \underline{37.77} & 36.44 & \underline{33.51} & \underline{35.91}
        & \textbf{51.00} & \underline{44.00} & 30.50 & \underline{38.00}
        & \underline{18.50} & 23.01 & \underline{30.82}
        & \underline{37.56}
        & \underline{34.76} \\
        
        SDPO
        & 31.35 & 32.66 & 27.71 & 30.57
        & 43.50 & 41.50 & 31.00 & 29.50
        & 8.00 & 22.15 & 25.73
        & 26.53
        & 27.61 \\
        
        RLSD
        & 34.19 & 32.63 & 31.13 & 32.65
        & 45.50 & 39.00 & \underline{33.50} & 31.00
        & 12.00 & 21.93 & 27.11
        & 33.30
        & 31.02 \\
        
        SDAR
        & 33.67 & \underline{37.03} & 31.95 & 34.22
        & 47.00 & 40.50 & 30.50 & 37.00
        & 16.00 & 22.80 & 29.07
        & 37.36
        & 33.55 \\
        SOD
        & 31.89 & 34.91 & 30.66 & 32.49
        & 38.50 & 34.00 & 28.00 & 24.50
        & 9.50 & 21.72 & 23.43
        & 22.28
        & 26.06 \\
        \midrule
        \textbf{\ourmodel(Ours)}
        & \textbf{44.61} & \textbf{40.16} & \textbf{38.44} & \textbf{41.00}
        & \underline{50.50} & \textbf{46.50} & \textbf{36.00} & \textbf{41.00}
        & \textbf{21.00} & \textbf{24.73} & \textbf{33.18}
        & \textbf{38.29}
        & \textbf{37.51} \\
        
        \hline
        \multicolumn{14}{c}{\rule[-0.7ex]{0pt}{3.2ex}\textit{\textbf{Qwen3-8B}}} \\
        \midrule
        Vanilla
        & 28.08 & 36.55 & 29.74 & 31.46
        & 46.50 & 49.50 & \underline{38.50} & 36.00
        & 10.50 & 21.29 & 29.26
        & 34.07
        & 31.60 \\
        
        GRPO
        & 35.97 & 39.78 & 35.02 & 36.92
        & 45.50 & 43.00 & 31.50 & 32.00
        & 8.50 & 22.15 & 26.66
        & 29.21
        & 30.93 \\
        
        ToolRL
        & 35.22 & 43.72 & 37.53 & 38.82
        & 56.00 & 49.00 & 33.00 & 35.50
        & 17.00 & 20.64 & 31.10
        & 35.68
        & 35.20 \\
        
        MatchTIR
        & \underline{40.88} & \underline{46.64} & \underline{40.83} & \underline{42.78}
        & \underline{57.00} & 49.00 & 37.00 & \underline{39.50}
        & \underline{20.00} & 23.87 & 33.78
        & 40.54
        & \underline{39.03} \\
        
        SDPO
        & 38.13 & 42.72 & 36.20 & 39.02
        & 43.00 & 41.00 & 33.50 & 32.00
        & 9.00 & 20.00 & 25.94
        & 35.21
        & 33.39 \\
        
        RLSD
        & 36.92 & 40.53 & 36.98 & 38.14
        & 46.50 & 45.50 & 33.00 & 36.00
        & 16.00 & \underline{26.02} & 30.63
        & 38.29
        & 35.69 \\
        
        SDAR
        & 35.02 & 44.23 & 36.72 & 38.66
        & 54.00 & \underline{51.00} & \underline{38.50} & 38.50
        & \underline{20.00} & 25.38 & \underline{34.09}
        & \underline{41.37}
        & 38.04 \\
        SOD
        & 30.08 & 40.92 & 32.58 & 34.53
        & 50.00 & 44.50 & 34.50 & 35.00
        & 13.00 & 19.14 & 28.54
        & 33.64
        & 32.23 \\
        \midrule
        \textbf{\ourmodel(Ours)}
        & \textbf{46.99} & \textbf{50.71} & \textbf{43.07} & \textbf{46.92}
        & \textbf{59.50} & \textbf{51.50} & \textbf{42.00} & \textbf{45.00}
        & \textbf{23.50} & \textbf{27.74} & \textbf{37.56}
        & \textbf{41.58}
        & \textbf{42.02} \\
        
        \bottomrule
    \end{tabular}
    }
    \caption{Main results of \ourmodel and baseline methods across three benchmarks using two LLM backbones. For BFCL, MF, MP, and LC denote the Miss Function, Miss Parameter, and Long Context subsets, respectively. The best and second-best results in each column are shown in bold and underlined, respectively.}
    \label{tab:main}
\end{table*}

\subsection{Policy Optimization}
\paratitle{Advantage Integration.}
To incorporate turn-level hindsight while preserving the original RL objective, we interpolate the base advantage with its hindsight-modulated counterpart. 
For a token at position $t$ belonging to turn $k$ within the $i$-th rollout, the integrated advantage is defined as
\[
\widetilde{A}_{i,t}
=
A_{i,t}^{\mathrm{base}}
\left[(1-\lambda)+\lambda w_{i,t}\right],
\]
where $\lambda$ controls the strength of hindsight modulation.

\paratitle{Objective Function.}
We retain the standard GRPO objective and replace the original advantage with $\widetilde{A}_{i,t}$:
\begin{equation*}
    \begin{split}
        \mathcal{J}_{\text{TurnSight}}(\theta)&=\mathbb{E}_{q\sim D,\{\tau_i\}\sim\pi_{\theta_\text{old}}(\cdot|q)}\frac{1}{G}\sum_{i=1}^{G}\frac{1}{|\tau_i|}\sum_{t=1}^{|\tau_i|}\bigg[\min(\\
        &\rho_{i,t}\widetilde{A}_{i,t},\text{clip}(\rho_{i,t},
        1-\epsilon,1+\epsilon)\widetilde{A}_{i,t})\bigg].
    \end{split}
\label{eq:turnsight}
\end{equation*}

\section{Experiments}
In this section, we first describe the experimental setups and then conduct an extensive evaluation of our \ourmodel.

\subsection{Experimental Setups}

\paratitle{Training Data.}
We use the FTRL dataset~\citep{ye2025ftrl} for all post-training experiments.
It provides more than 2,000 automatically generated problems paired with executable tool environments and programmatically verifiable feedback. 
The dataset covers single-step tool calls, sequential reasoning over multiple tools, and parallel tool invocations.

\paratitle{Evaluation Benchmarks.}
We evaluate \ourmodel on three benchmarks covering both in-domain and out-of-domain settings. FTRL serves as the in-domain benchmark, whereas BFCL~\citep{patil2025bfcl} and ToolHop~\citep{ye2025toolhop} are used for out-of-domain evaluation. Together, these benchmarks assess the model's ability to generalize beyond the task distributions and patterns encountered during training.

\paratitle{Baselines.}
To evaluate the effectiveness of \ourmodel,
we compare it with two families of baselines.
The first includes RL methods for TIR: \textbf{Vanilla}~\citep{yang2025qwen3}, \textbf{GRPO}~\citep{shao2024deepseekmath}, \textbf{ToolRL}~\citep{qian2025toolrl}, and \textbf{MatchTIR}~\citep{qu2026matchtir}, which provide supervision at the outcome, trajectory, and turn levels, respectively.
The second comprises representative self-distillation methods, including \textbf{SDPO}~\citep{hubotter2026sdpo}, \textbf{RLSD}~\citep{yang2026rlsd}, \textbf{SDAR}~\citep{lu2026sdar}, and \textbf{SOD}~\cite{zhong2026sod}, which leverage privileged contexts to guide RL optimization.

\paratitle{Implementation Details.}
We conduct experiments with Qwen3-4B and Qwen3-8B as the policy backbones \citep{yang2025qwen3}. All experiments are implemented with the verl framework~\citep{sheng2025hybridflow} and initialized directly from the corresponding base checkpoints, without an intermediate supervised
fine-tuning stage.
Each optimization batch contains 32 queries, with 16 trajectories sampled per query and at most 10 interaction turns per trajectory. 
We do not apply an explicit
KL-divergence regularizer to constrain the policy updates and train for 3 epochs on 8 NVIDIA A800 GPUs with 80~GB memory each. 
For \ourmodel, both the mixing coefficient $\lambda$ and the modulation bound $\epsilon_w$ are set to $0.5$. 
We construct 3 execution-conditioned teacher views using 3 lookahead depths. 
All teacher and reference branches are initialized
from the corresponding policy backbone and remain frozen throughout RL training.
Since RLSD and \ourmodel are agnostic to the underlying RL algorithm and can be combined with different policy-optimization methods, we use MatchTIR as the RL backbone for both.

\subsection{Experimental Results}

Table~\ref{tab:main} reports the results on three benchmarks, from which we make the following observations:

$\bullet$ Firstly, RL consistently improves tool-use performance, while outcome-only optimization generalizes poorly.
Compared with the vanilla Qwen3 models, all RL post-training methods achieve clear gains on the in-domain FTRL benchmark, demonstrating the effectiveness of RL for optimizing multi-turn tool-use behaviors. 
However, GRPO often exhibits noticeably weaker performance on the out-of-domain BFCL and ToolHop benchmarks, indicating that outcome-level rewards are insufficient for identifying which intermediate tool interactions should receive credit.

$\bullet$ Secondly, fine-grained credit assignment consistently leads to stronger and more transferable policies.
Compared with vanilla GRPO, methods equipped with structured tool-level supervision achieve consistently better performance across both model scales, confirming the importance of fine-grained credit signals for long-horizon TIR. 
The strong performance of MatchTIR further shows that accurate turn-level supervision is highly effective, making it a strong baseline for evaluating on-policy hindsight methods.

$\bullet$ Thirdly, simply combining on-policy self-distillation with RL is insufficient for effective TIR. 
Although SDPO, RLSD, and SDAR consistently outperform vanilla GRPO, they still lag behind methods with explicit turn-level credit assignment. 
This suggests that privileged information from ground-truth trajectories or successful rollouts cannot directly provide accurate supervision for every intermediate decision, especially in long-horizon tool interactions. Moreover, token-level self-distillation may introduce conflicting credit signals within the same interaction turn. These results highlight the importance of aggregating hindsight supervision at the turn level to provide reliable credit assignment.

$\bullet$ Finally, \ourmodel achieves the best overall performance while remaining fully on-policy.
Across the three benchmarks, \ourmodel consistently outperforms all baselines and establishes new state-of-the-art results on the 8B model, improving the previous best method by 7.7\% in overall average performance. 
Notably, the largest improvements are observed on challenging BFCL subsets such as Long Context and Miss Parameter, where successful tool use requires accurately attributing credit across multiple interaction turns. 
These results demonstrate that \ourmodel effectively combines the exploration benefits of on-policy RL with fine-grained hindsight supervision, enabling more accurate credit assignment without relying on reference trajectories.


\begin{table}[t]
    \centering
    \small
    \setlength{\tabcolsep}{3pt}
    \begin{tabular}{@{}lcccc@{}}
        \toprule
        Variant & Solve-P & Solve-R & Solve-F1 & Avg. \\
        \midrule
        \ourmodel & \textbf{46.99} & \textbf{50.71} & \textbf{43.07} & \textbf{46.92} \\
        \quad \wo turn-level aggregation & 41.61 & 47.53 & 40.54 & 43.23 \\
        \quad \wo group normalization  & 42.84 & 48.02 & 40.09 & 43.65 \\
        \quad \wo multi-teacher selection & 44.26 & 50.24 & 42.35 & 45.62 \\
        \bottomrule
    \end{tabular}
    \caption{Ablation study of the core components.}
    \label{tab:ablation}
\end{table}

\subsection{Ablation Study}
To quantify the contribution of each component, we conduct an ablation study on Qwen3-8B, with results reported in Table~\ref{tab:ablation}. Removing any component consistently degrades performance, confirming the significance of each element:

\begin{figure}[t]
    \centering
    \includegraphics[width=\linewidth]{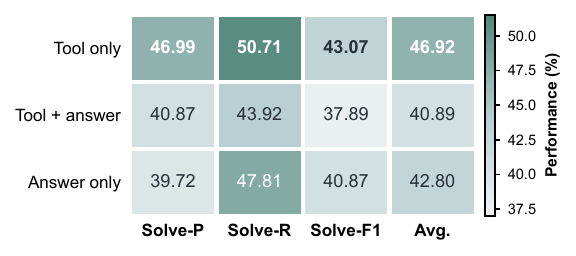}
    \caption{Effect of hindsight context composition. Darker cells indicate stronger performance.}
    \label{fig:context-composition}
\end{figure}

\begin{figure*}[t]
    \centering
    \includegraphics[width=\textwidth]{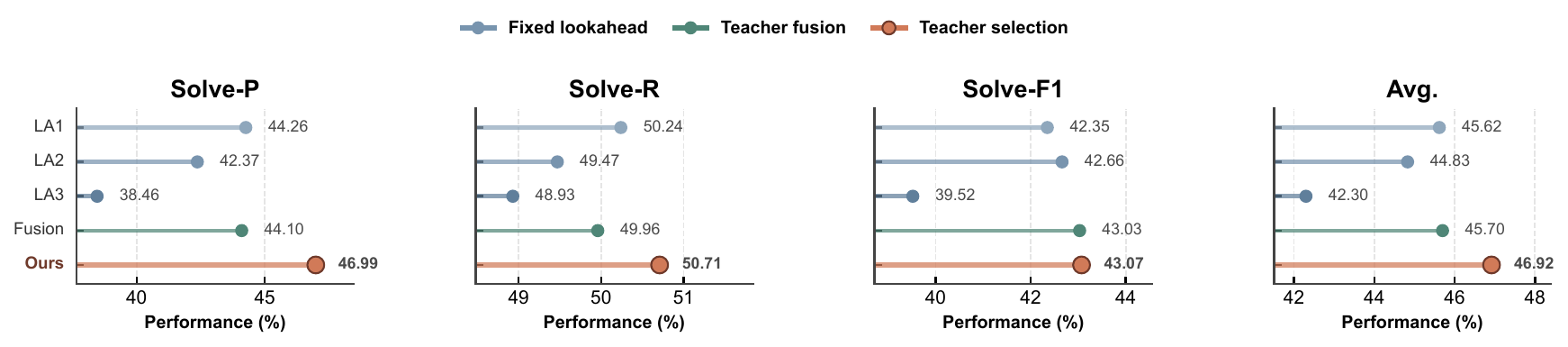}
    \caption{Comparison of fixed-lookahead teachers, teacher fusion, and our direction-consistent teacher selection.}
    \label{fig:lookahead-teacher}
\end{figure*}

\textbf{\wo turn-level aggregation} refers to a variant that removes turn-level aggregation and instead selects the best execution-conditioned hindsight view for each token independently, using token-level signals to modulate the RL advantage. 
The significant performance drop shows that multi-turn TIR is naturally organized around interaction turns rather than individual tokens. Applying supervision independently at the token level can introduce conflicting credit signals within the same tool interaction, whereas aggregating hindsight evidence at turn level produces more reliable credit assignment.

\textbf{\wo group normalization} is a variant that removes group normalization and directly uses the selected turn-level hindsight signal to modulate the RL advantage. 
The noticeable degradation demonstrates that the magnitude of hindsight signals varies substantially across rollout groups. Group normalization calibrates these signals into a comparable scale, leading to more stable and reliable credit assignment.

\textbf{\wo multi-teacher selection} denotes the variant that replaces the multi-lookahead teacher with a single execution-conditioned teacher using one fixed lookahead depth. 
The performance drop indicates that different lookahead depths provide complementary hindsight information: shorter horizons capture the immediate quality of a tool interaction, while longer horizons reveal delayed consequences that only become apparent after subsequent reasoning and tool use.

Overall, these results demonstrate that the effectiveness of \ourmodel comes from the coordinated design of turn-level aggregation, group normalization, and multi-lookahead supervision, which together produce hindsight signals that are more coherent, comparable, and temporally informative.

\subsection{Further Analysis}

\paratitle{Hindsight Context Composition.}
To evaluate which type of privileged information provides the most effective hindsight supervision, we compare three teacher context configurations: tool result only, tool result + ground-truth answer, and ground-truth answer only, while keeping all other settings unchanged.
As shown in Figure~\ref{fig:context-composition}, using only the tool result consistently achieves the best performance. 
Interestingly, adding the ground-truth answer further degrades performance, suggesting that more privileged information does not necessarily provide better supervision. 
We attribute this to the fact that tool results provide direct, state-aligned evidence for evaluating the current tool interaction, whereas the ground-truth answer contains trajectory-level information that may obscure turn-level causal contributions. 
These results highlight execution feedback as the most effective privileged signal for hindsight-based credit assignment.

\paratitle{Lookahead Teacher Selection.}
To evaluate the effect of hindsight horizon and teacher combination strategies, we compare fixed-lookahead teachers with depths of one, two, and three turns, a teacher-fusion variant, and our direction-consistent teacher selection strategy. 
As shown in Figure~\ref{fig:lookahead-teacher}, the one-step teacher performs best among the fixed-horizon variants, while longer lookahead depths consistently reduce performance, suggesting that excessive future context introduces irrelevant interactions that dilute turn-specific supervision. 
Although teacher fusion provides a slight improvement over the best fixed teacher, it cannot effectively resolve conflicting hindsight signals across different horizons. 
In contrast, our direction-consistent selection strategy achieves the best performance on all metrics, outperforming both the strongest fixed teacher and teacher fusion. 
These results demonstrate that multi-horizon hindsight is most effective when informative future evidence is selectively retained rather than indiscriminately aggregated.

\begin{figure}[t]
    \centering
    \includegraphics[width=\linewidth]{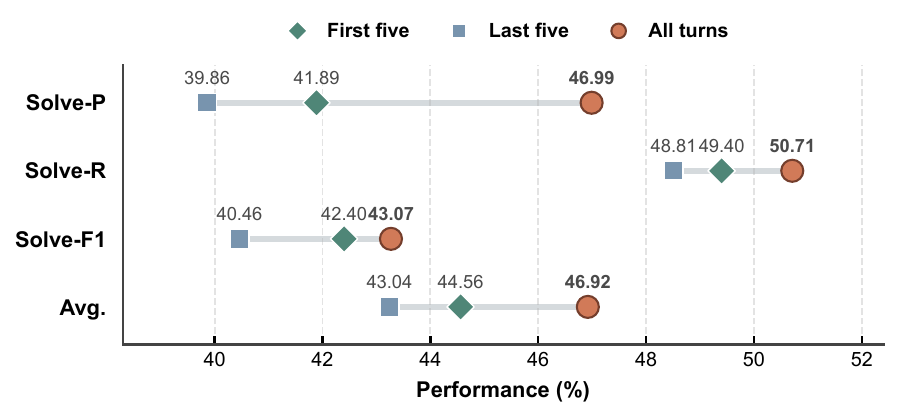}
    \caption{Effect of temporal self-distillation coverage.}
    \label{fig:temporal-coverage}
\end{figure}

\paratitle{Temporal Coverage of Hindsight.}
To evaluate how the temporal coverage of hindsight supervision affects multi-turn TIR, we compare three variants that apply self-distillation to the first five turns, last five turns, or all turns of each trajectory. 
As shown in Figure~\ref{fig:temporal-coverage}, supervising the entire trajectory consistently achieves the best performance, outperforming the first-five-turn and last-five-turn variants by 2.36\% and 3.88\% in average performance, respectively. 
Moreover, supervising the first five turns consistently outperforms the last five, suggesting that early tool-use decisions have a larger impact by shaping subsequent interaction states. 
Nevertheless, the clear advantage of full-trajectory supervision indicates that reliable credit assignment requires execution-conditioned hindsight throughout the entire interaction.

\begin{figure}[t]
    \centering
    \includegraphics[width=0.48\linewidth]{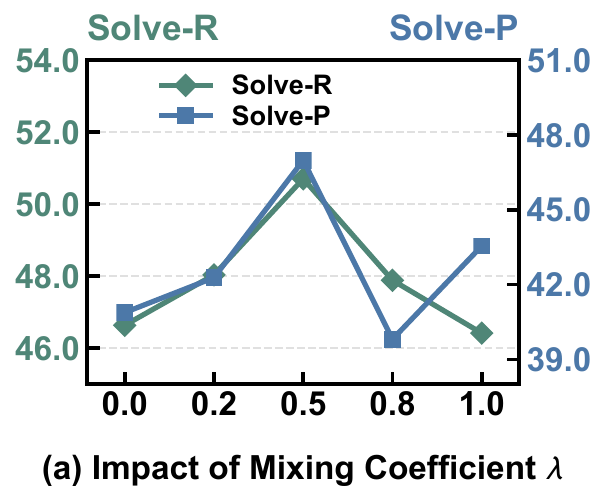}
    \hfill
    \includegraphics[width=0.48\linewidth]{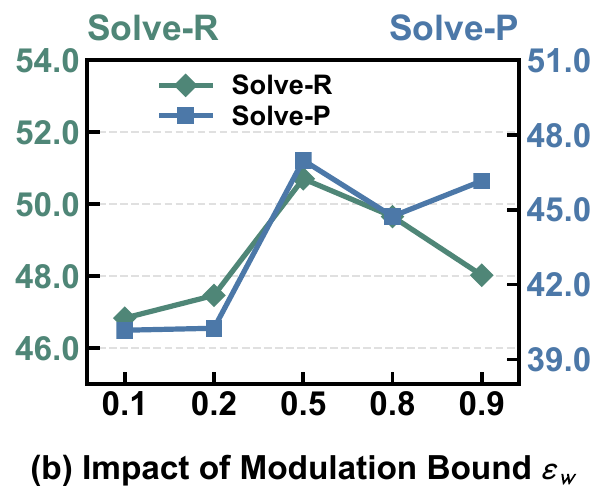}
    \caption{Sensitivity analysis of performance to hyper-parameters. (a) shows the impact of mixing coefficient $\lambda$. (b) illustrates the effect of modulation bound $\epsilon_w$.}
    \label{fig:analy_hyper}
\end{figure}

\paratitle{Hyper-parameter Analysis.} To examine the sensitivity of \ourmodel, we independently vary two key hyper-parameters, the mixing coefficient $\lambda$ and modulation bound $\epsilon_w$, and evaluate their effects on FTRL. 
As shown in Figure~\ref{fig:analy_hyper}, both exhibit a consistent unimodal trend, achieving the best performance at $\lambda=0.5$ and $\epsilon_w=0.5$.
Increasing $\lambda$ initially improves performance, indicating that turn-level hindsight signals complement trajectory-level advantages. However, an overly large $\lambda$ degrades performance, as excessive emphasis on local credit may weaken global task-level guidance. This highlights the need to balance fine-grained interaction supervision with overall outcome optimization. 
Similarly, $\epsilon_w$ determines the strength of hindsight modulation. Small values restrict the contribution of execution-conditioned signals, whereas overly large values may amplify noise in imperfect teacher--student gaps. The optimal intermediate setting therefore strikes an effective balance between exploiting hindsight information and maintaining stable RL optimization.

\section{Conclusion}
In this paper, we identify two key requirements for credit assignment in TIR: alignment with on-policy execution states and coherence at the interaction level. 
To address this, we propose \ourmodel, a turn-level hindsight self-distillation framework that derives execution-conditioned, multi-horizon hindsight assessments from on-policy tool execution outcomes. 
By leveraging these assessments to modulate RL advantages without altering their optimization direction, \ourmodel enables fine-grained and reliable credit assignment while preserving the exploration capability of RL. 
Extensive experiments on both in-domain and out-of-domain benchmarks demonstrate the effectiveness, robustness, and generalization ability of \ourmodel across diverse multi-turn TIR scenarios.

\bibliography{aaai2027}

@misc{shao2024deepseekmath,
  title         = {{DeepSeekMath}: Pushing the Limits of Mathematical Reasoning in Open Language Models},
  author        = {Shao, Zhihong and Wang, Peiyi and Zhu, Qihao and Xu, Runxin and Song, Junxiao and Bi, Xiao and Zhang, Haowei and Zhang, Mingchuan and Li, Y. K. and Wu, Y. and Guo, Daya},
  year          = {2024},
  eprint        = {2402.03300},
  archivePrefix = {arXiv},
  primaryClass  = {cs.CL}
}

@inproceedings{
    hubotter2026sdpo,
    title={Reinforcement Learning via Self-Distillation},
    author={Jonas H{\"u}botter and Frederike L{\"u}beck and Lejs Deen Behric and Anton Baumann and Marco Bagatella and Daniel Marta and Ido Hakimi and Idan Shenfeld and Thomas Kleine Buening and Carlos Guestrin and Andreas Krause},
    booktitle={Forty-third International Conference on Machine Learning},
    year={2026},
    url={https://openreview.net/forum?id=QkfkxyRizZ}
}

@inproceedings{
    shenfeld2026sdft,
    title={Self-Distillation Enables Continual Learning},
    author={Idan Shenfeld and Mehul Damani and Jonas H{\"u}botter and Pulkit Agrawal},
    booktitle={Forty-third International Conference on Machine Learning},
    year={2026},
    url={https://openreview.net/forum?id=qA6FgH0nnZ}
}

@inproceedings{
    zhao2026opsd,
    title={Self-Distilled Reasoner: On-Policy Self-Distillation for Large Language Models},
    author={Siyan Zhao and Zhihui Xie and Mengchen Liu and Jing Huang and Guan Pang and Feiyu Chen and Aditya Grover},
    booktitle={Forty-third International Conference on Machine Learning},
    year={2026},
    url={https://openreview.net/forum?id=Jpxfof0EaS}
}

@article{ye2026opcd,
  title={On-policy context distillation for language models},
  author={Ye, Tianzhu and Dong, Li and Wu, Xun and Huang, Shaohan and Wei, Furu},
  journal={arXiv preprint arXiv:2602.12275},
  year={2026}
}

@article{li2026rethinking,
  title={Rethinking on-policy distillation of large language models: Phenomenology, mechanism, and recipe},
  author={Li, Yaxuan and Zuo, Yuxin and He, Bingxiang and Zhang, Jinqian and Xiao, Chaojun and Qian, Cheng and Yu, Tianyu and Gao, Huan-ang and Yang, Wenkai and Liu, Zhiyuan and others},
  journal={arXiv preprint arXiv:2604.13016},
  year={2026}
}

@article{kim2026does,
  title={Why Does Self-Distillation (Sometimes) Degrade the Reasoning Capability of LLMs?},
  author={Kim, Jeonghye and Luo, Xufang and Kim, Minbeom and Lee, Sangmook and Kim, Dohyung and Jeon, Jiwon and Li, Dongsheng and Yang, Yuqing},
  journal={arXiv preprint arXiv:2603.24472},
  year={2026}
}

@misc{yang2026rlsd,
  title         = {Self-Distilled {RLVR}},
  author        = {Yang, Chenxu and Qin, Chuanyu and Si, Qingyi and Chen, Minghui and Gu, Naibin and Yao, Dingyu and Lin, Zheng and Wang, Weiping and Wang, Jiaqi and Duan, Nan},
  year          = {2026},
  eprint        = {2604.03128},
  archivePrefix = {arXiv},
  primaryClass  = {cs.LG}
}

@misc{lu2026sdar,
  title         = {Self-Distilled Agentic Reinforcement Learning},
  author        = {Lu, Zhengxi and Yao, Zhiyuan and Han, Zhuowen and Wang, Zi-Han and Wu, Jinyang and Gu, Qi and Cai, Xunliang and Lu, Weiming and Xiao, Jun and Zhuang, Yueting and Shen, Yongliang},
  year          = {2026},
  eprint        = {2605.15155},
  archivePrefix = {arXiv},
  primaryClass  = {cs.LG}
}

@misc{ma2026sdsearch,
  title         = {{SD-Search}: On-Policy Hindsight Self-Distillation for Search-Augmented Reasoning},
  author        = {Ma, Yufei and Liang, Zihan and Chen, Ben and Qian, Zhipeng and Dai, Huangyu and Mao, Lingtao and Zhang, Xuxin and Lei, Chenyi and Ou, Wenwu},
  year          = {2026},
  eprint        = {2605.18299},
  archivePrefix = {arXiv},
  primaryClass  = {cs.AI}
}

@misc{zhang2026stepopsd,
  title         = {{StepOPSD}: Step-Aware Online Preference Distillation for Agent Reinforcement Learning},
  author        = {Zhang, Yanfei and Lin, Xu and Wu, Chenglin},
  year          = {2026},
  eprint        = {2605.27140},
  archivePrefix = {arXiv},
  primaryClass  = {cs.AI}
}

@article{ding2026sgcd,
  title={Keep Policy Gradient in Charge: Sibling-Guided Credit Distillation for Long-Horizon Tool-Use Agents},
  author={Ding, Tianyu and Xin, Jianhong and Weinstein, Juan Pablo De la Cruz},
  journal={arXiv preprint arXiv:2606.12634},
  year={2026}
}

@misc{zhou2026sageopd,
  title         = {{SAGE-OPD}: Selective Agent-Guided Intervention for Multi-Turn On-Policy Distillation},
  author        = {Zhou, Yuhang and Zhang, Lizhu and Wu, Yifan and Wang, Mingyi and Peng, Bo and Liu, Jiayi and Fan, Xiangjun and Zhao, Zhuokai},
  year          = {2026},
  eprint        = {2606.19659},
  archivePrefix = {arXiv},
  primaryClass  = {cs.CL}
}

@misc{zhou2026turnopd,
  title         = {{TurnOPD}: Making On-Policy Distillation Turn-Aware for Efficient Long-Horizon Agent Training},
  author        = {Zhou, Yuhang and Zheng, Kai and Li, Haoling and Peng, Dengyun and Xu, Can and Chen, Jingjing},
  year          = {2026},
  eprint        = {2607.05804},
  archivePrefix = {arXiv},
  primaryClass  = {cs.AI}
}

@inproceedings{qu2026matchtir,
  title={MatchTIR: Fine-Grained Supervision for Tool-Integrated Reasoning via Bipartite Matching},
  author={Qu, Changle and Dai, Sunhao and Cai, Hengyi and Xu, Jun and Wang, Shuaiqiang and Yin, Dawei},
  booktitle={Proceedings of the 64th Annual Meeting of the Association for Computational Linguistics (Volume 1: Long Papers)},
  year={2026}
}

@inproceedings{ye2025toolhop,
  title     = {{ToolHop}: A Query-Driven Benchmark for Evaluating Large Language Models in Multi-Hop Tool Use},
  author    = {Ye, Junjie and Du, Zhengyin and Yao, Xuesong and Lin, Weijian and Xu, Yufei and Chen, Zehui and Wang, Zaiyuan and Zhu, Sining and Xi, Zhiheng and Yuan, Siyu and others},
  booktitle = {Proceedings of the 63rd Annual Meeting of the Association for Computational Linguistics},
  pages     = {2995--3021},
  year      = {2025}
}

@misc{yang2025qwen3,
  title         = {{Qwen3} Technical Report},
  author        = {Yang, An and Li, Anfeng and Yang, Baosong and Zhang, Beichen and Hui, Binyuan and Zheng, Bo and Yu, Bowen and Gao, Chang and Huang, Chengen and Lv, Chenxu and others},
  year          = {2025},
  eprint        = {2505.09388},
  archivePrefix = {arXiv},
  primaryClass  = {cs.CL}
}

@inproceedings{sheng2025hybridflow,
  title     = {{HybridFlow}: A Flexible and Efficient {RLHF} Framework},
  author    = {Sheng, Guangming and Zhang, Chi and Ye, Zilingfeng and Wu, Xibin and Zhang, Wang and Zhang, Ru and Peng, Yanghua and Lin, Haibin and Wu, Chuan},
  booktitle = {Proceedings of the Twentieth European Conference on Computer Systems},
  pages     = {1279--1297},
  year      = {2025}
}

@article{gou2023tora,
  title={Tora: A tool-integrated reasoning agent for mathematical problem solving},
  author={Gou, Zhibin and Shao, Zhihong and Gong, Yeyun and Yang, Yujiu and Huang, Minlie and Duan, Nan and Chen, Weizhu and others},
  journal={In Proceedings of the 12th International Conference on Learning Representations (ICLR)},
  year={2024}
}

@article{qu2025tool,
  title={Tool learning with large language models: A survey},
  author={Qu, Changle and Dai, Sunhao and Wei, Xiaochi and Cai, Hengyi and Wang, Shuaiqiang and Yin, Dawei and Xu, Jun and Wen, Ji-Rong},
  journal={Frontiers of Computer Science},
  volume={19},
  number={8},
  pages={198343},
  year={2025},
  publisher={Springer}
}

@article{li2025torl,
  title={Torl: Scaling tool-integrated rl},
  author={Li, Xuefeng and Zou, Haoyang and Liu, Pengfei},
  journal={arXiv preprint arXiv:2503.23383},
  year={2025}
}

@inproceedings{
    qian2025toolrl,
    title={Tool{RL}: Reward is All Tool Learning Needs},
    author={Cheng Qian and Emre Can Acikgoz and Qi He and Hongru WANG and Xiusi Chen and Dilek Hakkani-T{\"u}r and Gokhan Tur and Heng Ji},
    booktitle={The Thirty-ninth Annual Conference on Neural Information Processing Systems},
    year={2025},
    url={https://openreview.net/forum?id=eOLdGbXT6t}
}

@article{feng2025retool,
  title={Retool: Reinforcement learning for strategic tool use in llms},
  author={Feng, Jiazhan and Huang, Shijue and Qu, Xingwei and Zhang, Ge and Qin, Yujia and Zhong, Baoquan and Jiang, Chengquan and Chi, Jinxin and Zhong, Wanjun},
  journal={arXiv preprint arXiv:2504.11536},
  year={2025}
}

@article{wang2025otc,
  title={Otc: Optimal tool calls via reinforcement learning},
  author={Wang, Hongru and Qian, Cheng and Zhong, Wanjun and Chen, Xiusi and Qiu, Jiahao and Huang, Shijue and Jin, Bowen and Wang, Mengdi and Wong, Kam-Fai and Ji, Heng},
  journal={arXiv e-prints},
  pages={arXiv--2504},
  year={2025}
}

@article{zhang2025nemotron,
  title={Nemotron-research-tool-n1: Tool-using language models with reinforced reasoning},
  author={Zhang, Shaokun and Dong, Yi and Zhang, Jieyu and Kautz, Jan and Catanzaro, Bryan and Tao, Andrew and Wu, Qingyun and Yu, Zhiding and Liu, Guilin},
  journal={arXiv preprint arXiv:2505.00024},
  year={2025}
}

@inproceedings{
    xue2025simpletir,
    title={Simple{TIR}: End-to-End Reinforcement Learning for Multi-Turn Tool-Integrated Reasoning},
    author={Zhenghai Xue and Longtao Zheng and Qian Liu and Yingru Li and Xiaosen Zheng and Zejun MA and Bo An},
    booktitle={The Fourteenth International Conference on Learning Representations},
    year={2026},
    url={https://openreview.net/forum?id=EplNy91Xqh}
}

@inproceedings{zeng2025tool,
  title={Tool Zero: Training Tool-Augmented LLMs via Pure RL from Scratch},
  author={Zeng, Yirong and Ding, Xiao and Hou, Yutai and Wang, Yuxian and Du, Li and Dai, Juyi and Ding, Qiuyang and Tang, Duyu and Tu, Dandan and Liu, Weiwen and others},
  booktitle={Findings of the Association for Computational Linguistics: EMNLP 2025},
  pages={9135--9147},
  year={2025}
}

@article{wei2025autotir,
  title={Autotir: Autonomous tools integrated reasoning via reinforcement learning},
  author={Wei, Yifan and Yu, Xiaoyan and Weng, Yixuan and Pan, Tengfei and Li, Angsheng and Du, Li},
  journal={arXiv preprint arXiv:2507.21836},
  year={2025}
}

@inproceedings{
    qu2025from,
    title={From Exploration to Mastery: Enabling {LLM}s to Master Tools via Self-Driven Interactions},
    author={Changle Qu and Sunhao Dai and Xiaochi Wei and Hengyi Cai and Shuaiqiang Wang and Dawei Yin and Jun Xu and Ji-Rong Wen},
    booktitle={The Thirteenth International Conference on Learning Representations},
    year={2025},
    url={https://openreview.net/forum?id=QKBu1BOAwd}
}

@inproceedings{
    chang2025thor,
    title={{THOR}: Tool-Integrated Hierarchical Optimization via {RL} for Mathematical Reasoning},
    author={Qikai Chang and Zhenrong Zhang and Pengfei Hu and Jun Du and Jiefeng Ma and Yicheng Pan and Jianshu Zhang and Quan Liu and Jianqing Gao},
    booktitle={The Fourteenth International Conference on Learning Representations},
    year={2026},
    url={https://openreview.net/forum?id=0Af7UiJISU}
}

@article{
    jiang2025verltool,
    title={VerlTool: Towards Holistic Agentic Reinforcement Learning with Tool Use},
    author={Dongfu Jiang and Yi Lu and Zhuofeng Li and Zhiheng Lyu and Ping Nie and Haozhe Wang and Alex Su and Hui Chen and Kai Zou and Chao Du and Tianyu Pang and Wenhu Chen},
    journal={Transactions on Machine Learning Research},
    issn={2835-8856},
    year={2026},
    url={https://openreview.net/forum?id=g2LCOW43Md},
    note={}
}

@inproceedings{
    wang2025information,
    title={Information Gain-based Policy Optimization: A Simple and Effective Approach for Multi-Turn Search Agents},
    author={Guoqing Wang and Sunhao Dai and Guangze Ye and Zeyu Gan and Wei Yao and Yong Deng and Xiaofeng Wu and Zhenzhe Ying},
    booktitle={The Fourteenth International Conference on Learning Representations},
    year={2026},
    url={https://openreview.net/forum?id=qkWP6phrvZ}
}

@inproceedings{ye2025ftrl,
  author       = {Junjie Ye and
                  Changhao Jiang and
                  Zhengyin Du and
                  Yufei Xu and
                  Xuesong Yao and
                  Zhiheng Xi and
                  Xiaoran Fan and
                  Qi Zhang and
                  Tao Gui and
                  Xuanjing Huang and
                  Jiecao Chen},
  title        = {Feedback-Driven Tool-Use Improvements in Large Language Models via
                  Automated Build Environments},
  booktitle    = {Findings of the Association for Computational Linguistics, {ACL} 2026,
                  San Diego, California, United States, July 2-7, 2026},
  pages        = {2293--2323},
  publisher    = {Association for Computational Linguistics},
  year         = {2026}
}

@inproceedings{patil2025bfcl,
title={The Berkeley Function Calling Leaderboard (BFCL): From Tool Use to Agentic Evaluation of Large Language Models}, 
author={Patil, Shishir G. and Mao, Huanzhi and Cheng-Jie Ji, Charlie and Yan, Fanjia and Suresh, Vishnu and Stoica, Ion and E. Gonzalez, Joseph},
booktitle={Forty-second International Conference on Machine Learning},
year={2025},
}

@inproceedings{
    yu2025dapo,
    title={{DAPO}: An Open-Source {LLM} Reinforcement Learning System at Scale},
    author={Qiying Yu and Zheng Zhang and Ruofei Zhu and Yufeng Yuan and Xiaochen Zuo and YuYue and Weinan Dai and Tiantian Fan and Gaohong Liu and Juncai Liu and LingJun Liu and Xin Liu and Haibin Lin and Zhiqi Lin and Bole Ma and Guangming Sheng and Yuxuan Tong and Chi Zhang and Mofan Zhang and Ru Zhang and Wang Zhang and Hang Zhu and Jinhua Zhu and Jiaze Chen and Jiangjie Chen and Chengyi Wang and Hongli Yu and Yuxuan Song and Xiangpeng Wei and Hao Zhou and Jingjing Liu and Wei-Ying Ma and Ya-Qin Zhang and Lin Yan and Yonghui Wu and Mingxuan Wang},
    booktitle={The Thirty-ninth Annual Conference on Neural Information Processing Systems},
    year={2025},
}

@article{zhong2026sod,
  title={Sod: Step-wise on-policy distillation for small language model agents},
  author={Zhong, Qiyong and Zheng, Mao and Song, Mingyang and Lin, Xin and Sun, Jie and Jiang, Houcheng and Wang, Xiang and Fang, Junfeng},
  journal={arXiv preprint arXiv:2605.07725},
  year={2026}
}

@article{sang2026crisp,
  title={Crisp: Compressed reasoning via iterative self-policy distillation},
  author={Sang, Hejian and Xu, Yuanda and Zhou, Zhengze and He, Ran and Wang, Zhipeng and Sun, Jiachen},
  journal={arXiv preprint arXiv:2603.05433},
  year={2026}
}

@article{li2026unifying,
  title={Unifying group-relative and self-distillation policy optimization via sample routing},
  author={Li, Gengsheng and Yang, Tianyu and Fang, Junfeng and Song, Mingyang and Zheng, Mao and Guo, Haiyun and Zhang, Dan and Wang, Jinqiao and Chua, Tat-Seng},
  journal={arXiv preprint arXiv:2604.02288},
  year={2026}
}

@article{wang2026skill,
  title={Skill-sd: Skill-conditioned self-distillation for multi-turn llm agents},
  author={Wang, Hao and Wang, Guozhi and Xiao, Han and Zhou, Yufeng and Pan, Yue and Wang, Jichao and Xu, Ke and Wen, Yafei and Ruan, Xiaohu and Chen, Xiaoxin and others},
  journal={arXiv preprint arXiv:2604.10674},
  year={2026}
}

\clearpage
\appendix
\section*{Appendix}

\section{Benchmark and Evaluation Details}
\label{sec:appendix_benchmarks}

We use FTRL for post-training and in-domain evaluation, while BFCL and ToolHop assess generalization to unseen tool interfaces and interaction patterns. Table~\ref{tab:appendix_dataset_statistics} summarizes the resulting data composition. The reported counts refer to executable queries or task instances.

\begin{table*}[t]
    \centering
    \small
    \setlength{\tabcolsep}{6pt}
    \begin{tabular}{llp{0.52\textwidth}r}
        \toprule
        \textbf{Benchmark} & \textbf{Usage} & \textbf{Split or Subset} & \textbf{Instances} \\
        \midrule
        FTRL & Post-training & All training environments & 2,215 \\
        FTRL & In-domain evaluation & Single-Hop / Parallel Single-Hop / Multi-Hop / Parallel Multi-Hop & 50 / 50 / 50 / 50 \\
        BFCL & Out-of-domain evaluation & Base / Missing Function / Missing Parameter / Long Context & 200 / 200 / 200 / 200 \\
        BFCL & Out-of-domain evaluation & Web Search / Memory & 200 / 465 \\
        ToolHop & Out-of-domain evaluation & Multi-hop tool-use queries & 995 \\
        \bottomrule
    \end{tabular}
    \caption{Data used for post-training and evaluation. Instance counts are reported for each benchmark split or subset.}
    \label{tab:appendix_dataset_statistics}
\end{table*}

\paratitle{FTRL.}\footnote{\url{https://github.com/bytedance/FTRL}}
FTRL~\citep{ye2025ftrl} provides executable tool-use tasks with automatically constructed queries, local tool environments, and programmatically verifiable feedback. We use all 2,215 training instances for post-training and evaluate on four test categories, each containing 50 queries. These categories differ in their dependency structures: Single-Hop tasks require a single tool-mediated step; Parallel Single-Hop tasks contain independent requests that can be resolved without cross-step dependencies; Multi-Hop tasks require later actions to use earlier execution results; and Parallel Multi-Hop tasks combine independent branches with sequential dependencies.

FTRL evaluates tool-use behavior at the subtask level through Solve-P, Solve-R, and Solve-F1. Let $N_{\mathrm{call}}$ denote the number of generated tool calls, $N_{\mathrm{solved}}$ the number of successfully resolved subtasks, and $N_{\mathrm{req}}$ the total number of required subtasks. The metrics are defined as
\begin{equation}
\mathrm{Solve\text{-}P}=
\begin{cases}
N_{\mathrm{solved}}/N_{\mathrm{call}}, & N_{\mathrm{call}}>0,\\
1, & N_{\mathrm{call}}=0.
\end{cases}
\end{equation}
\begin{equation}
\mathrm{Solve\text{-}R}
=
N_{\mathrm{solved}}/N_{\mathrm{req}}.
\end{equation}
and
\begin{equation}
\mathrm{Solve\text{-}F1}
=
\frac{2\,\mathrm{Solve\text{-}P}\,\mathrm{Solve\text{-}R}}
{\mathrm{Solve\text{-}P}+\mathrm{Solve\text{-}R}}.
\end{equation}
Because post-training and evaluation use the same environment family, FTRL measures in-domain improvements in executable multi-turn tool reasoning.

\paratitle{BFCL.}\footnote{\url{https://github.com/ShishirPatil/gorilla/tree/main/berkeley-function-call-leaderboard}}
BFCL~\citep{patil2025bfcl} evaluates function-calling and agentic behavior under diverse interaction conditions. Our evaluation includes four multi-turn subsets and two agentic subsets. The multi-turn portion comprises {Base}, {Missing Function}, {Missing Parameter}, and {Long Context}. These subsets respectively assess stateful function calling, behavior under unavailable tools, behavior under incomplete arguments, and tool selection from lengthy specifications containing distracting information. The agentic portion includes {Web Search}, which requires external information acquisition, and {Memory}, which evaluates the reuse of information obtained earlier in an interaction.

For multi-turn tasks, correctness is determined by the resulting environment state and, when applicable, the validity of the response path. State-modifying requests are checked through backend-state verification, whereas read-oriented requests must also satisfy the benchmark's response requirements. The agentic subsets use exact matching on the designated answer field. We report the official accuracy of each subset together with their average.

\paratitle{ToolHop.}\footnote{\url{https://huggingface.co/datasets/bytedance-research/ToolHop}}
ToolHop~\citep{ye2025toolhop} contains 995 multi-hop queries and 3,912 locally executable tools. Its tasks require agents to compose dependent tool calls over a tool collection that differs from the FTRL training environment. We therefore use ToolHop to examine whether the learned credit-assignment strategy transfers to new tools and interaction structures. Performance is measured by final-answer accuracy, with a prediction counted as correct only when its answer matches the verified target after the required tool interactions.

\section{Additional Implementation Details}
\label{sec:appendix_implementation}

All experiments are implemented with the verl framework~\citep{sheng2025hybridflow}. Table~\ref{tab:appendix_hyperparameters} reports the shared training and rollout configuration for Qwen3-4B and Qwen3-8B. Both models are initialized from their corresponding base checkpoints and trained for three epochs without an explicit KL penalty.

\begin{table}[t]
    \centering
    \small
    \setlength{\tabcolsep}{5pt}
    \begin{tabular}{lc}
        \toprule
        \textbf{Configuration} & \textbf{Value} \\
        \midrule
        Query batch size & 32 \\
        Validation batch size & 256 \\
        Maximum prompt length & 10,000 tokens \\
        Maximum response length & 13,000 tokens \\
        Learning rate & $1\times10^{-6}$ \\
        PPO mini-batch size & 32 \\
        Explicit KL coefficient & 0 \\
        Rollout engine & vLLM \\
        Sampling temperature & 1.0 \\
        Rollout GPU utilization & 0.7 \\
        Rollouts per query & 16 \\
        Maximum interaction turns & 10 \\
        Training epochs & 3 \\
        Checkpoint interval & 5 iterations \\
        Evaluation interval & 5 iterations \\
        Training hardware & $8\times$ A800-80GB \\
        \midrule
        Lookahead depths $\mathcal{D}_H$ & $\{1,2,3\}$ \\
        Mixing coefficient $\lambda$ & 0.5 \\
        Modulation bound $\epsilon_w$ & 0.5 \\
        \bottomrule
    \end{tabular}
    \caption{Training, rollout, and method-specific hyperparameters shared across experiments.}
    \label{tab:appendix_hyperparameters}
\end{table}

\paratitle{Prompt-Group Normalization.}
For each query, we form one normalization pool from the selected hindsight gaps at every valid policy-token position across all 16 sampled rollouts. A single mean and standard deviation are computed from this pool and shared by all turns in the prompt group; trajectories are not partitioned according to turn index. Invalid or masked positions, including environment-generated observation tokens, do not contribute to these statistics. This implementation allows trajectories of different lengths to be normalized together without introducing artificial values for turns they do not contain.

\paratitle{Multi-Horizon Teacher Evaluation.}
The hindsight teachers share a frozen reference policy but differ in the amount of future execution information included in their contexts. Specifically, we construct teacher views with lookahead depths $1$, $2$, and $3$. Each teacher scores the student's sampled response through teacher forcing, after which the token-level gaps are aggregated into turn-level hindsight assessments. The resulting statistics, teacher-selection outputs, and advantage-modulation weights are detached from the computational graph. Environment-generated observations are excluded from the policy loss because they are not actions sampled by the agent. All privileged contexts and teacher computations are used only during training; deployment requires only the learned policy.

\paratitle{Evaluation Protocol.}
All compared methods use the same benchmark environments and decoding settings. For the BFCL Web Search subset, search results are obtained through the Google Search endpoint of the Serper API.\footnote{\url{https://serper.dev/}} The remaining BFCL subsets, together with FTRL and ToolHop, are evaluated in their released local execution environments. This shared protocol ensures that differences in performance are attributable to the learned policies rather than changes in tool execution or evaluation conditions.


\end{document}